\documentclass[preprint,12pt]{elsarticle}

\usepackage{comment}

\usepackage{amssymb}
\usepackage{amsthm}
\usepackage{amsmath}
\usepackage{bm}
\usepackage{subcaption}

\usepackage[figuresright]{rotating}
\usepackage{graphicx}
\usepackage{amsmath}
\usepackage{amssymb}

\begin{document}

\begin{frontmatter}




\title{Spatial Partial Functionalization of Neural Networks\\ based on Noise Fields}





\author[address1]{Shuhei Ikemoto\corref{mycorrespondingauthor}}
\cortext[mycorrespondingauthor]{Corresponding author}
\ead{s.ikemoto@ieee.org}
\ead[url]{http://www.brain.kyutech.ac.jp/~ikemoto/}

\author[address2]{Fabio DallaLibera}
\ead{fabiodl@gmail.com}

\address[address1]{Graduate School of Life Science and Systems Engineering, Kyushu Institute of Technology, 2-4 Hibikino, Wakamatsu, Kitakyushu, Fukuoka, 8080196, Japan}
\address[address2]{Graduate School of Engineering Science, Osaka University, 1-3 Machikaneyama, Toyonaka, Osaka, Japan}

\begin{abstract}

Noise in neural computation is typically regarded as a disturbance, but its spatial distribution may also actively regulate which parts of a network participate in computation. 
This paper investigates the spatial partial functionalization of Noise-modulated Neural Networks using noise fields. 
We first present an activation function suitable for this goal, the crossing activation function, using the sample-level, statistical-level, and analytical-level implementations, and examine parameter reuse across these implementations. 
We then introduce a virtual noise field, an auxiliary continuous space for generating spatially structured network noise fields that activate partially overlapping subnetworks. 
Using one-dimensional function approximation tasks, we evaluate how multiple functions can be stored in a single network when each function is assigned to a different noise-field location. 
The results show that memory capacity improves when the spatial arrangement of noise fields reflects the proximity relationships among the functions to be learned, whereas mismatches in noise field structure can reduce effective capacity. 
These findings suggest that structured noise can serve not only as a perturbation but also as a topology-defining factor for functional subnetwork selection.

\end{abstract}


\end{frontmatter}

\section{Introduction}
\label{sec:intro}

Neural activity in the brain is highly heterogeneous, exhibiting significant spatial and temporal variability \cite{Raichle2006brain,Logothetis2008what}.
Functional brain imaging and electrophysiological measurements have repeatedly demonstrated that, depending on specific tasks or contexts, only limited regions within the brain become selectively active.
However, the criteria and mechanisms that regulate which specific neural circuits or regions transition into an active state in a task-specific manner remain poorly understood.
Although it is thought that various factors such as synaptic connectivity\cite{Katz1996synaptic,Luo2021architectures}, neurotransmitters\cite{Bargmann2012beyond,Marder2012neuromodulation}, and history dependence\cite{Buonomano2009state,Morcos2016} act as internal states, no unified principle has been established to explain them.

Alongside such brain activity, fluctuations inherent in neural activity, namely neural noise\cite{destexhe2012neuronal,birn2012role}, are constantly observed.
Neural noise has been studied for many years, and it has been demonstrated that its diverse generation mechanisms\cite{faisal2008noise} and its intensity are closely related to the state of neural activity and task performance\cite{mcintosh2008increased,garrett2012modulation}.
In particular, there is a strong tendency for the variance and variability of neural responses to increase in regions where activity is heightened, and it is almost taken for granted that there is a strong correlation between the locality of noise intensity and the locality of brain activity\cite{garrett2012modulation,Nomi2017moment}.
However, these reports implicitly assume a causal relationship in which changes in brain activity lead to changes in noise characteristics.
To our knowledge, no studies have reversed this perspective to examine the possibility that the spatial distribution of noise intensity determines the temporal and spatial heterogeneity of brain activity.

While noise is always detrimental in linear systems, in non-linear systems, an appropriate noise level improves system performance. Indeed, stochastic resonance refers to a phenomenon in which the loss of information throughput that occurs in nonlinear systems is mitigated by the addition of noise with an appropriate intensity\cite{gammaitoni1998stochastic,wiesenfeld1995stochastic}.
Although it was initially proposed as a mathematical model to explain the periodic occurrence of glacial and interglacial periods\cite{benzi1981mechanism,benzi1982stochastic}, it has since been universally observed in a wide range of systems, including electronic circuits\cite{fauve1983stochastic,mcnamara1988observation}, mechanical systems\cite{landa2000vibrational}, and chemical reaction systems\cite{guderian1996stochastic}.
In particular, within biological nervous systems, stochastic resonance is regarded as one of the principles that actively utilizes inevitable noise\cite{mcdonnell2011benefits,moss2004stochastic}, and its role has been suggested and observed at various levels, such as sensory reception\cite{douglass1993noise,levin1996broadband}, neural information processing\cite{mcdonnell2009stochastic}, and motor control\cite{priplata202noise,martinez2007stochastic}.
Furthermore, to explain and reproduce its properties, various mathematical models have been proposed, including models based on bistable systems\cite{fauve1983stochastic}, stochastic differential equations\cite{jung1991amplification}, and discrete state transition models\cite{mcnamara1989theory,zozor1999stochastic}.
Consequently, the concept that noise is not merely a source of error but an element that plays a functional role is now widely accepted in the biological nervous system.

Spiking neural networks (SNNs), often regarded as the third generation of neural network models, provide a computational framework in which information is represented by discrete spiking events rather than continuous firing-rate variables \cite{maass1997networks,ponulak2011introduction,tavanaei2019deep}.
Because spike generation is intrinsically nonlinear and event-based, SNNs have also served as an important testbed for stochastic resonance, where appropriately tuned noise can transform subthreshold inputs into reliable spike timing or signal propagation \cite{ozer2009stochastic,ozer2010weak}.
These studies suggest that noise can be treated not merely as an external perturbation to neural computation but as a parameter that regulates whether, when, and through which pathways neural events are generated.
In this field, we have proposed a mathematical model that not only utilizes noise but also incorporates it to such an extent that the model cannot function without it\cite{ikemoto2018noise,ikemoto2021noise}. This model, named the Noise-modulated Neural Network, enables training and inference only when and where noise is applied.
Specifically, in this network, hidden layer units to which noise is not applied always output zero, regardless of the input signal, and information from the input signal is reflected probabilistically in the output signal only when noise is applied.
This feature makes it possible to apply noise to only a portion of the entire network and to use only that subnetwork for training and inference.
In fact, we have previously reported that by partitioning the regions where noise is applied, this network can learn and utilize multiple functions without failure\cite{ikemoto2021noise}.
The partial application of noise in the Noise-modulated Neural Network and the resulting partial utilization of the network demonstrate that in studies of stochastic resonance, not only the intensity of the noise but also the perspective of "where noise is applied" can serve as a parameter.
Although highly simplified, the proposed model shows that the temporal and spatial heterogeneity of activity, and specifically of noise, commonly observed in biological brains, is a key component for making a neural network learn and perform inference on multiple functions. In other terms, studying the effects of the application of a "noise field", or more intuitively, "where and how much noise is applied", is of interest both for research on stochastic resonance and as a learning model for brain mechanisms.

In this study, we expand upon this concept by formalizing the theoretical foundations of the Noise-modulated Neural Network and clarifying its relationship with deterministic neural networks. 
We will then investigate in detail the memory capacity, that is, how well multiple functions can be stored when utilizing subnetworks designated by multiple noise fields.
In regard to the theoretical foundations, we investigate three types of computational implementations: A) Sample-level implementation: using the events themselves generated by actually adding noise, B) Statistical-level implementation: using statistics of the events generated by actually adding noise, and C) Analytical-level implementation: using theoretically equivalent deterministic functions.
Through the analysis of these three implementations, we clarify their similarities and differences.
In regard to the memory capacity, by continuously varying the “noise field,” we change the subnetworks used while allowing for overlapping regions. 
By repeatedly performing tests to learn different functions, we quantitatively demonstrate that the similarity of the learned functions and the configuration of the subnetwork strongly influences memory capacity. 

The main contributions of this paper are summarized as follows:
\begin{enumerate}
    \item Formulated noise-modulated neural networks using sample-level, statistical-level, and analytical-level implementations, and clarified their relationships in terms of parameter reuse.
    \item Proposed a virtual noise field as a representation for generating a structured noise field that activates a subnetwork.
    \item Demonstrated that the memory capacity of a single network is enhanced when the spatial arrangement of the noise field reflects the proximity relationships among the functions being learned.
\end{enumerate}

This paper is organized as follows. 
Section~\ref{sec:nn} formalizes the Noise-modulated Neural Network by introducing the crossing activation function, its Sample-level, Statistical-level, and Analytical-level implementations, and the noise field as a topology-defining hyperparameter. 
Section~\ref{sec:parameter} examines the relationship among these three implementations through parameter-transfer experiments and evaluates whether trainable parameters can be reused across implementation levels. 
Section~\ref{sec:multi-reg} introduces the virtual noise field, a spatial representation used to generate structured network noise fields, and demonstrates how different localized noise patterns can selectively functionalize different subnetworks within a single model. 
Section~\ref{sec:memory} evaluates the memory capacity of the proposed framework by assigning multiple target functions to different positions in the virtual noise field and analyzing how function similarity, noise field dimensionality, and spatial arrangement affect simultaneous learning. 
Section~\ref{sec:discussion} discusses the characteristic loss behavior observed in the memory-capacity experiments, focusing on the effects of grid resolution and dimensional imbalance in the virtual noise field. 
Finally, Section~\ref{sec:conclusion} concludes the paper by summarizing the findings and outlining future directions.

\section{Noise-modulated Neural Network}
\label{sec:nn}

\begin{figure}[t!]
  \centering
  \includegraphics[width=0.8\linewidth]{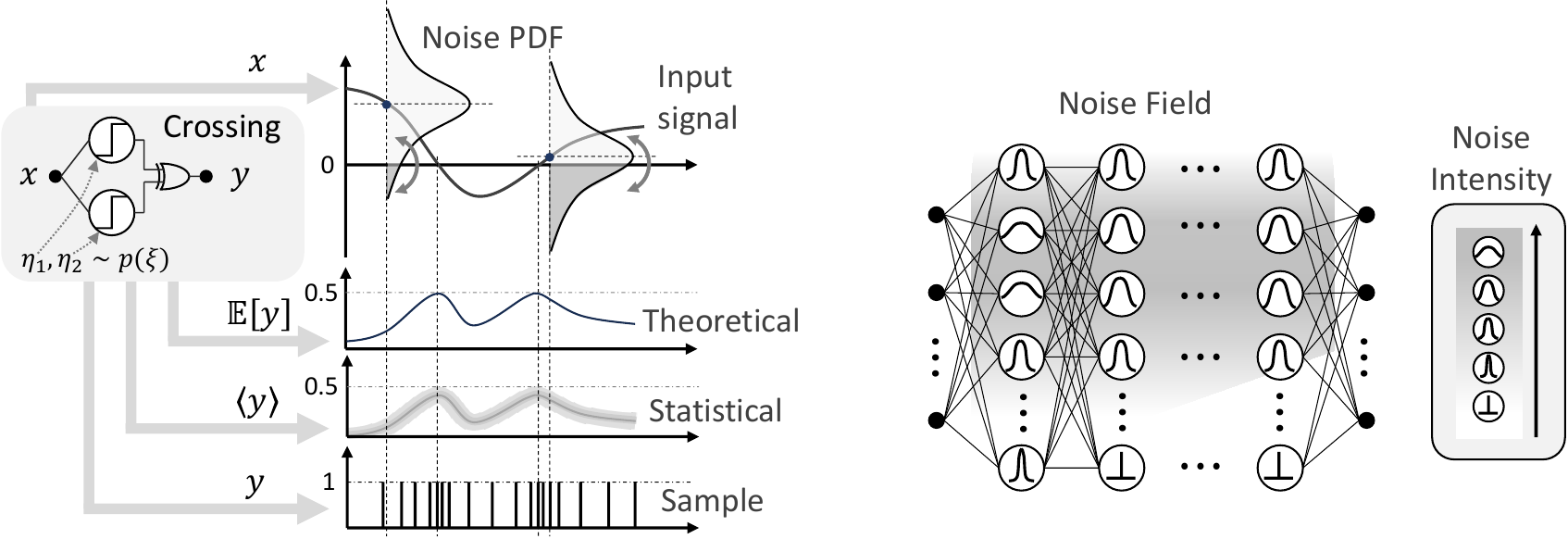}
  \caption{Overview of Noise-modulated Neural Network.
  Left: the crossing activation function and its three different implementation levels. The Analytical-level, Statistical-level, and Sample-level behave as a deterministic continuous function, a stochastic continuous function, and a binary function, depending on whether the noise is used in the probability density distribution, a sample group, or an individual sample, respectively.
  Right: the noise field on the noise-modulated neural network using the crossing activation function. 
  Noise for each activation function is given as a hyperparameter over the network.}
  \label{fig:overview}
\end{figure}

Figure~\ref{fig:overview} shows an overview of the key features in the noise-modulated neural network originally proposed in \cite{ikemoto2021noise}.
This section formally defines this model to clarify its functionalities and features in detail.
In Sec.~\ref{subsec:cross}, we introduce the crossing activation function, which is the source of this model's functionality, and its three levels of implementation.
Based on the formal definition, we explain parameter sharing and reuse in Sec.~\ref{subsec:network}.
Then, in Sec.~\ref{subsec:noisefield}, we introduce the noise field as the main focus of this paper.

\subsection{Crossing Activation Function}
\label{subsec:cross}

To disable learning and inference in the absence of noise, the ``crossing'' activation function has been proposed in \cite{ikemoto2021noise}. This is a type of spiking model that outputs 1 at the moment when the threshold is crossed and 0 otherwise. Our activation function is defined as follows:
\begin{eqnarray}
  \label{eq:cr_orig}
  z = \phi(d) =
   \begin{cases}
    1 & if ~ d \ge \eta_1 ~ \dot{\lor} ~ d \ge \eta_2\\
    0 & otherwise,
   \end{cases}
\end{eqnarray}
where $d \in \mathbb{R}$ is an input value, $\dot{\lor}$ is the logic operator XOR, and $\eta_1, \eta_2 \overset{\text\small\textrm{iid}}{\sim} p(\xi)$ are noise values.
Intuitively, this activation function represents a simple spiking neuron model that fires when the input signal crosses zero due to noise.
It is clear that if noise is absent, the crossing event must not occur for a constant input $d$.

Defining the cumulative probability of the noise probability density function $p(\xi)$ as:
\begin{equation}
  \label{eq:th_expected}
  F(d) = P(d \ge \eta) = \int^{d}_{-\infty} p(\xi) d\xi,
\end{equation}
the expected value of the crossing activation function's output can be written as follows:
\begin{equation}
  \label{eq:cr}
  \mathbb{E}[z] = \bar{\phi}(d) = 2 F(d)(1 - F(d)).
\end{equation}


By definition, the cumulative probability $F: \mathbb{R}\to[0, 1]$ is a monotonically non-decreasing function. Consequently, the mapping $\bar{\phi}:\mathbb{R}\to[0,0.5]$ naturally forms a unimodal, bell-shaped curve—functionally analogous to a radial basis function—peaking at $0.5$ when $F(d) = 0.5$, provided $F(d)$ does not plateau over an intermediate interval.

Considering the $\bar{\phi}(\cdot)$ as the statistically equivalent activation function of the original crossing function in Eq.\ref{eq:cr_orig}, its derivative is given by the following:
\begin{equation}
  \label{eq:cr'}
  \bar{\phi}(d)' = 2 (1-2F(d))p(d)
\end{equation}
where $p(d)$ is the value of the noise probability density function $p(\xi)$ evaluated at $\xi=d$. 

Suppose that $p(\xi)$ is currently unknown, but we can obtain $N$ samples $\eta_i \sim p(\xi)$. 
In this case, $F(d)$ can be approximated by the empirical cumulative distribution function $F_N(d)$, defined as follows:
\begin{equation}
  \label{eq:th_mean}
  F_N(d) = \frac{1}{N} \sum_{i=1}^{N} \mathbf{1}_{\{d \ge \eta_i\}} \approx F(d)
\end{equation}
where $N$ indicates the total number of samples taken from $p(\xi)$, and $\mathbf{1}_{\{\cdot\}}$ is the indicator function, which evaluates to $1$ if the condition $d \ge \eta_i$ is true and $0$ otherwise. Intuitively, $F_N(d)$ represents the sample mean of these indicator values.
Regarding $p(d)$, it can be approximated using kernel density estimation as follows:
\begin{equation}
  \label{eq:kde}
  p(d) \approx \frac{1}{nh} \sum_{i=1}^{N} K\left( \frac{d-\eta_i}{h} \right)
\end{equation}
where $K$ and $h$ indicate a kernel function and a bandwidth.
In particular, when a uniform kernel is used, it can be written as follows:
\begin{equation}
  \label{eq:kde_u}
  p(d) \approx \frac{1}{2h} \left( F_N^{+}(d)-F_N^{-}(d) \right)
\end{equation}
where $F_N^{+}(d)$ and $F_N^{-}(d)$ denote approximate values of $F_N(d)$ obtained by modifying the condition $d \ge \eta_i$ in Eq.\ref{eq:th_mean} to $d+h \ge \eta_i$ and $d-h \ge \eta_i$, respectively. In a similar manner,  let us denote by
$z^+_i \in \{ 1,0 \}$ and $z^-_i \in \{ 1,0 \}$ the values of the samples of $z$ obtained by substituting $d+h$ or $d-h$ for $d$ in the condition of Eq.\ref{eq:cr_orig}, respectively, and by $\bar{z}^+$ and $\bar{z}^-$ their respective means. The  approximations for $\bar{\phi}(d)$ and $\bar{\phi}(d)'$ can be obtained as follows:

\begin{eqnarray}
  \label{eq:app_cr}
  \phi_N(d) &=& \frac{1}{2} \left( \bar{z}^+ + \bar{z}^- \right) \approx \bar{\phi}(d)\\
  \label{eq:app_cr'}
  \phi_N(d)' &=& \frac{1}{2h} \left( \bar{z}^+ - \bar{z}^- \right) \approx \bar{\phi}(d)'
\end{eqnarray}
Note that as $h$ becomes smaller, the approximation error decreases, but so does the difference between $\bar{z}^+$ and $\bar{z}^-$.
By definition, the following relationships hold:
\begin{eqnarray}
  \phi_N(d) &=& \frac{1}{N} \sum_{i=1}^{N} \frac{1}{2} \left( z^+_i + z^-_i \right)\\
  \phi_N(d)' &=& \frac{1}{N} \sum_{i=1}^{N} \frac{1}{2h} \left( z^+_i - z^-_i \right)
\end{eqnarray}
From these relationships, the following function, which corresponds to the original activation function $\phi$, can also be derived:
\begin{eqnarray}
  \label{eq:sample_cr}
  \phi_1(d) &=& \frac{1}{2} \left( z^+_1 + z^-_1 \right) \xrightarrow{h \to 0} \phi(d)\\
  \label{eq:sample_cr'}
  \phi_1(d)' &=& \frac{1}{2h} \left( z^+_1 - z^-_1 \right) \xrightarrow{h \to 0} \delta(d)
\end{eqnarray}
where $\delta(d)$ represents the Dirac delta function. 
Because $z^+$ and $z^-$ are binary, the $\phi_1(\cdot)$ and $\phi_1(\cdot)'$ are ternary-valued functions.

In summary, the crossing activation function can be considered at the following three levels:
\begin{itemize}
    \item \textbf{Sample-level}: Use $\phi$ or $\phi_1$ as an activation function and $\phi_1'$ as its derivative.
    \item \textbf{Statistical-level}: Use $\phi_N$ as an activation function and $\phi_N'$ as its derivative.
    \item \textbf{Analytical-level}: Use $\bar{\phi}$ as an activation function and $\bar{\phi}'$ as its derivative.
\end{itemize}
Note that the Sample-level outputs a stochastic spike-like signal, the Statistical-level outputs a stochastic continuous value, and the Analytical-level outputs a deterministic continuous value.
Although there are significant differences between these levels, they are all derived from the same equation Eq.\ref{eq:cr_orig}, and their behavior depends on the value of the noise and its statistical properties.

The above formulation is related to existing approaches for training networks with non-differentiable or stochastic units, such as straight-through estimators and surrogate-gradient methods \cite{bengio2013estimating,courbariaux2016binarized,lee2016training,neftci2019surrogate}. 
These methods typically replace the derivative of a discontinuous operation with a heuristic or smooth surrogate during backpropagation. 
In contrast, the present formulation uses noise not only to define a backward gradient but also to reshape the forward computation itself. 
The statistical properties of the injected noise, therefore, provide a common basis for both the forward response and its backward derivative, preserving consistency between inference and learning. 
This perspective is closely aligned with the concept of positive noise learning, in which noise is treated as a functional component of computation rather than as a disturbance.

\subsection{Network Implementation}
\label{subsec:network}

Consider an $L$-layer feed-forward Noise-modulated Neural Network.
Let $\mathbf{z}^{(0)} = \mathbf{x}$ denote the input to the network.
The forward pass through the hidden layers $l = 1, \dots, L-1$ is given by
\begin{eqnarray}
  \label{eq:nnn1}
  \mathbf{d}^{(l)} &=& \mathbf{W}^{(l)} \mathbf{z}^{(l-1)} + \mathbf{b}^{(l)}, \\
  \label{eq:nnn2}
  \mathbf{z}^{(l)} &=& \Phi^{(l)}(\mathbf{d}^{(l)}),
\end{eqnarray}
where $\mathbf{W}^{(l)}$ and $\mathbf{b}^{(l)}$ are the weight matrix and bias vector of layer $l$, and $\Phi^{(l)}$ denotes the element-wise application of the crossing activation function (whose precise form depends on the level of implementation mentioned above).
The output layer applies only a linear transformation without activation:
\begin{equation}
  \label{eq:nnn3}
  \mathbf{y} = \mathbf{W}^{(L)} \mathbf{z}^{(L-1)}.
\end{equation}
The set of trainable parameters is
\begin{equation}
  \label{eq:nnn4}
  \mathbf{\theta} = \{ \mathbf{W}^{(l)}, \mathbf{b}^{(l)} \}_{l=1}^{L-1} \cup \{ \mathbf{W}^{(L)} \}.
\end{equation}
It is important to note that, unlike other approaches like the parametric bias~\cite{rybicki2009reinforcement}, the activation layers $\Phi^{(l)}$ contain no trainable parameters; noise for each activation function is given as a hyperparameter, and all learnable quantities reside exclusively in the weight matrices and bias vectors.
Therefore, whether $\mathbf{\theta}$ can be shared or reused across the aforementioned three levels of crossing activation functions depends on whether their behavior is the same or similar when used in the network.

Although the Statistical-level and the Analytical-level are defined differently, they represent the same deterministic input-output mapping in the limit of large sample sizes. 
In the Statistical-level implementation, the activation function $\phi_N$ and its derivative $\phi_N'$ are obtained by sample-based approximations of $\bar{\phi}$ and $\bar{\phi}'$, respectively. 
Therefore, for a fixed input value $d$, we have
\begin{equation}
    \phi_N(d) \xrightarrow{N \to \infty} \bar{\phi}(d), \qquad
    \phi_N'(d) \xrightarrow{N \to \infty,\ h \to 0} \bar{\phi}'(d).
\end{equation}
Consequently, the network implemented at the Statistical-level converges to the Analytical-level network, and the same set of trainable parameters $\theta$ can be shared between these two levels, up to the finite-sample approximation error.

In contrast, the Sample-level implementation is not simply a finite-sample approximation of the Analytical-level network when multiple nonlinear layers are stacked. 
At the Sample-level, the output of each activation layer is a stochastic binary or ternary variable. 
Therefore, the input of the $l$-th layer $d^{(l)}$ is itself a random variable due to the stochasticity of $z^{(l-1)}$.
Denoting the expected value of $d^{(l)}$ as $\bar{d}^{(l)}$ and assuming that $h$ is sufficiently small, the following relationship holds:
\begin{equation}
    \mathbb{E}\left[ \Phi_1^{(l)} \left( \mathbf{d}^{(l)} \right) \right] \neq \mathbb{E}\left[ \Phi_1^{(l)} \left( \bar{\mathbf{d}}^{(l)} \right) \right] = \bar{\Phi}^{(l)} \left( \bar{\mathbf{d}}^{(l)} \right)
\end{equation}
because the distributions of what we are taking the expected value of are different.
In other words, the $\Phi_1$ carry information not only through their mean firing probabilities but also through their full distributions, as opposed to $\Phi_N$ and $\bar{\Phi}$.
For this reason, even if the Sample-level output is averaged over many noise realizations, the resulting expected network output is not generally identical to the output of the Statistical-level or Analytical-level networks. 
Specifically, because the expectation operator and the nonlinear activation function do not commute, substituting the expected input into the activation function does not yield the expected output. 

On the other hand, the Analytical-level and its finite-sample approximation, namely the Statistical-level, can be seen as a mean-field approximation of the Sample-level network.
The difference manifests specifically within the hidden layers of multi-layer architectures. 
Rather than propagating the full joint probability distribution of the stochastically firing neurons, the mean-field approach propagates only their expected values, thereby discarding the effects of variance and inter-neuron correlations. 
Naturally, this discrepancy does not exist in a single-layer network receiving purely deterministic inputs, but in deep networks, the approximation differences can accumulate across successive layers.
Therefore, parameters trained at the Statistical-level or Analytical-level cannot, in general, be directly reused at the Sample-level without additional calibration or fine-tuning. 
They can, however, constitute a useful starting point, as will be experimentally shown later.

\subsection{Noise Field}
\label{subsec:noisefield}

Although not explicitly mentioned thus far, the behavior of the crossing activation function varies depending on the noise applied internally, regardless of the three implementation levels discussed earlier. 
In other words, just like the number of layers, the number of hidden units in each layer, and connectivity, additive noise is one of the hyperparameters of noise-modulated neural networks.
However, compared to other hyperparameters, there is a neurophysiological justification for modifying additive noise during or between the inference and learning processes.

Let $p^{(l,k)}$ denote the probability density function of the noise applied within the crossing activation function of the $k$-th unit in the $l$-th layer.
The probability density distribution of the noise can be specified independently for each layer and unit.
In other words, the hyperparameter governing noise represents the spatial distribution to which units in which layers are subjected to noise and the probability density distribution with which they are subjected.
Therefore, we refer to this hyperparameter as the “noise field” and define it as follows:
\begin{equation}
\mathcal{P} = \bigcup_{l=1}^{L-1} \mathcal{P}^{(l)} = \left \{ p^{(1,1)}, p^{(1,2)}, \dots, p^{(2,1)}, p^{(2,2)}, \dots \right \}.
\end{equation}
If $p^{(l,k)} = \delta$ holds, i.e. the noise value for the $k$-th unit on the $l$-th layer is deterministically zero, the output and the derivative are always zero, regardless of the input and the implementation level, as explained in Sec.\ref{subsec:cross}.
This neuron is, in all respects, uninfluential on the network output. Therefore, the modulation of each crossing activation function by the noise field can also produce discrete structural changes, such as gating at the unit level.

The primary aim of introducing the noise field is to equip the Noise-modulated Neural Network with a form of controllable network topology.
As shown above, in an extreme case, applying noise to only a subset of units activates the corresponding subnetwork for learning and inference, while the units left without noise remain inert and are effectively detached from the computation.
By assigning different noise fields, one can therefore select different subnetworks within the same set of trainable parameters, and by switching the noise field repeatedly, multiple functions can be learned in a single network while suppressing mutual interference. 
In this sense, the noise field acts as a topology-defining hyperparameter that determines which portion of the network is recruited for each task.

\section{Parameter reuse across different algorithm implementations}
\label{sec:parameter}

In this section, we empirically validate the efficacy of parameter sharing and reuse among the Analytical-level, the Statistical-level, and the Sample-level implementations introduced in Sec.~\ref{subsec:cross}.
Throughout this experiment, we compare three networks that share the same architecture but implement the crossing activation function at three different levels.
Each network is a feed-forward network with two hidden layers of 100 units each and identical crossing activation functions, differing only in the implementation level of those activations. 
Following the terminology of Sec.~\ref{subsec:cross}, we refer to them as the Analytical-level, Statistical-level, and Sample-level networks. The Analytical-level network evaluates the deterministic expected activation $\bar{\phi}$ in the closed form of Eq.~\ref{eq:analytic_cdf} and therefore requires no noise sampling. The Statistical-level network instead uses the sample-based activation $\phi_N$ and its derivative $\phi_N'$, where the $N$ is set to $10$ for the Statistical-level activation function.
The Sample-level network, denoted Sample 100 in the following, has the same architecture but implements each activation at the Sample-level (Eq.~\ref{eq:sample_cr}), and its output is obtained by averaging the network response over 100 independent noise realizations.

Furthermore, throughout the remainder of this paper, we assume that the noise follows a normal distribution $p(\xi) = \mathcal{N}(0, \sigma^2)$.
This assumption allows the $P(d \ge \eta)$ in Eq.~\ref{eq:cr} and Eq.~\ref{eq:cr'} to be written in the following form:
\begin{equation}
  \label{eq:analytic_cdf}
  P(d \ge \eta) = F_G\!\left(\frac{d}{\sigma}\right) = \frac{1}{2}\left[1 + \mathrm{erf}\!\left(\frac{d}{\sigma\sqrt{2}}\right)\right]
\end{equation}
where $F_G(\cdot)$ denotes the cumulative distribution function of the standard normal distribution. Substituting it as $F(d)$ in Eq. (3) gives the analytical-level activation function used in the following experiments.
It is clear that this implementation is entirely deterministic and requires no noise sampling. 
It represents the exact expected behavior of the Statistical-level implementation in the limit of infinite samples ($N \to \infty$).

To verify the sharing and reuse of trainable parameters across the different computational implementations of the Noise-modulated Neural Network, we conduct a 1D curve-fitting experiment.
The target function is defined as $y = \sin(x)$, evaluated over the domain $x \in [-2\pi, 2\pi]$. The dataset consists of $1000$ uniformly spaced data points.
The training pipeline is structured to evaluate both independent learning capabilities and cross-implementation parameter transfer.
For all training procedures, we utilize the Adam optimizer with a learning rate of $3 \times 10^{-4}$, zero weight decay, and Mean Squared Error (MSE) as the objective loss function.

The experiment is divided into two phases.
In the first phase, all models are initialized independently and trained for $50,000$ epochs. 
During this full learning process, a snapshot of the fully connected layer weights is captured every $100$ epochs. 
This establishes a baseline for the convergence properties of each individual implementation and generates a repository of intermediate parameter states.
In the second phase, to quantitatively evaluate parameter compatibility and reuse across different implementation levels, we perform a parameter transfer experiment.
At specific training milestones, denoted as switch locations $E_{\text{switch}} \in \{1000, 10000, 25000\}$ epochs, we extract the learned parameters $\theta$ from a trained source model. 
These weights are then injected into a destination model of a different implementation type. 
The destination model resumes training from this transferred state, continuing until the total number of epochs reaches $30,000$ (i.e., training for an additional $30,000 - E_{\text{switch}}$ epochs). 
The resulting convergence rates and final losses are evaluated against baseline models trained entirely from scratch, allowing us to ascertain the degree to which parameters learned in one noise regime can be generalized to another.

Table~\ref{tab:weightswap} reports destination-model performance under both non-switch and switch training conditions, with losses evaluated on the same regression task. 
The three destination formulations considered are Analytical, Statistical, and Sample 100. 
The row labeled Untrained corresponds to standard end-to-end training from random initialization (equivalent to switch location 0) and therefore provides the reference trajectory for each destination model. 
The remaining rows correspond to weight-swap experiments at predefined switch epochs, where the source representation (Analytical, Statistical, or Sample 100) is injected into the destination model, and optimization is continued thereafter. 
For each destination, two values are reported: Init (loss at the switching instant, i.e., the first loss after loading the switched state) and Final (loss at the end of post-switch training). 
These metrics allow for direct evaluation of both the immediate transfer quality at handoff (Init) and the asymptotic convergence after adaptation (Final).
Relative to Untrained, a lower Init loss indicates a more compatible transferred representation at the moment of switch, while a lower Final indicates superior asymptotic adaptation under that transfer condition. Cases where source and destination coincide serve as within-method continuity controls, whereas cross-method rows quantify cross-representation transferability. Together, the Init-to-Final gap provides an estimate of post-switch plasticity: large reductions indicate substantial recoverable mismatch after transfer, while small gaps suggest either immediate compatibility or limited residual adaptation. 
These results show that, in all transfer cases, pretrained weights reduce the initial loss after switching compared with random initialization, and no clear evidence of convergence to detrimental local minima is observed after post-switch optimization.

In summary, these results indicate that trainable parameters are reusable across the three implementations as effective initial conditions. 
Parameters trained under one implementation, whether Analytical-level, Statistical-level, or Sample-level, consistently reduced the loss at the switching point when transferred to another implementation. 
After post-switch optimization, the destination models reached comparable final losses without clear evidence of detrimental local minima. 
These results suggest that, although the three implementations are not strictly equivalent, they share a compatible parameter space for practical initialization and transfer.

\begin{table}[]
    \centering
\begin{tabular}{lcccccc}
\hline
Condition & \multicolumn{2}{c}{Analytical} & \multicolumn{2}{c}{Statistical} & \multicolumn{2}{c}{Sample 100} \\
 & Init & Final & Init & Final & Init & Final \\
\hline
Untrained & 5.14e-01 & 3.80e-08 & 4.99e-01 & 1.94e-02 & 5.49e-01 & 3.66e-03 \\
Switch 1000 / Analytical & 1.67e-02 & 5.91e-08 & 1.14e-01 & 2.03e-02 & 4.74e-02 & 3.63e-03 \\
Switch 1000 / Statistical & 3.82e-02 & 1.10e-07 & 7.35e-02 & 2.05e-02 & 3.96e-02 & 3.70e-03 \\
Switch 1000 / Sample 100 & 3.03e-02 & 4.15e-07 & 6.77e-02 & 2.08e-02 & 2.82e-02 & 3.56e-03 \\
Switch 10000 / Analytical & 9.19e-07 & 1.25e-06 & 2.11e-01 & 2.05e-02 & 9.79e-02 & 3.65e-03 \\
Switch 10000 / Statistical & 4.84e-03 & 5.01e-08 & 2.50e-02 & 2.08e-02 & 1.05e-02 & 3.65e-03 \\
Switch 10000 / Sample 100 & 1.10e-02 & 2.76e-08 & 5.09e-02 & 1.92e-02 & 5.78e-03 & 3.55e-03 \\
Switch 25000 / Analytical & 1.02e-07 & 1.67e-07 & 2.20e-01 & 3.20e-02 & 1.10e-01 & 7.84e-03 \\
Switch 25000 / Statistical & 4.24e-03 & 6.24e-05 & 2.01e-02 & 1.97e-02 & 8.86e-03 & 3.28e-03 \\
Switch 25000 / Sample 100 & 2.29e-02 & 5.32e-05 & 5.56e-02 & 2.02e-02 & 3.85e-03 & 3.42e-03 \\
\hline
\end{tabular}    
    \caption{Experimental evaluation of cross-implementation parameter transferability. The table reports the Mean Squared Error (MSE) at the exact moment of weight transfer (Init) and after subsequent post-switch optimization (Final). The 'Untrained' row provides the baseline convergence trajectory from random initialization.}
    \label{tab:weightswap}
\end{table}

\section{Spatial partial training and inference}
\label{sec:multi-reg}

In this section, we formalize the spatial partial functionalization of Noise-modulated Neural Networks by spatially localized noise patterns.
The noise field introduced in Section ~\ref{subsec:noisefield} considers independent noise for each neuron.
While neurons in standard artificial neural networks are often depicted as spatially organized arrays, traditional models lack any intrinsic definition of physical proximity or topological structure among units.
However, for brain-inspired models, it is desirable to utilize a noise field that reflects the spatial localization of noise intensity based on the spatial positions of the neurons.
To introduce such a constraint, this section introduces an abstract representation (referred to as a virtual noise field) that implements the noise field in Section ~\ref{subsec:noisefield} (referred to as a network noise field) with a spatial structure.
Specifically, the virtual noise field is defined as an auxiliary $V$-dimensional continuous space $[0,1]^V$, which introduces the property that “nearby locations receive noise of similar intensity” within this space.
We realize this spatial and partial functionalization by establishing rules to transform it into the network noise field.
In the following, we introduce a framework for noise field generation in Section ~\ref{subsec:noisefieldgeneration}, and in Section ~\ref{subsec:resultmultifunc}, we demonstrate the validity of this framework and provide examples of its application.
In addition, based on the results shown in Section~\ref{sec:parameter}, we use only the Analytical-level network hereafter.

\subsection{Noise Field Generation}
\label{subsec:noisefieldgeneration}

The key idea of the virtual noise field is to place a truncated Gaussian inside a $V$-dimensional unit hypercube $[0,1]^V$ to introduce the relationship that similarly intense noise is added to nearby locations.
The parameter $V \in \mathbb{N}$ represents the dimension of the proximity relationship; the larger the value, the more spatially complex the patterns generated.
This space is not the neural network itself; it is an auxiliary continuous space whose sole purpose is to generate spatially structured noise patterns.
The network noise field is generated through a unique mapping from the virtual noise field.
By changing the location of the truncated Gaussian center in the virtual noise field, we obtain different network noise field patterns.
If there is an overlapping active region between patterns in the virtual noise field, corresponding patterns in the network noise field also share overlapping active regions.

Truncated Gaussian centers are placed on a regular grid, the index set of which is defined as follows:
\begin{equation}
    \label{eq:Kv}
    \mathcal{K}_v = \prod_{i=1}^{V} \left\{ 1, \dots,m_v^{(i)} \right\}
\end{equation}
where $m_v^{(i)}$ is the number of grid points along the $i$-th axis.
The total number of grid points is therefore:
\begin{equation}
    \label{eq:Nv}
    N_v = |\mathcal{K}_v| = \prod_{i=1}^{V} m_v^{(i)}.
\end{equation}
A grid index $\mathbf{k}_v \in \mathcal{K}_v$ is converted to a continuous center position in $[0,1]^V$, whose $i$-th axis is:
\begin{equation}
\label{eq:rawcenter}
c_{v}^{(i)}=
\begin{cases}
\dfrac{k_{v}^{(i)}-1}{m_{v}^{(i)}-1} & \text{if } m_{v}^{(i)}>1,\\[4pt]
0 & \text{if } m_{v}^{(i)}=1.
\end{cases}
\end{equation}
To prevent the Gaussian from being cut off at the boundaries, we map these raw coordinates inward using the following correction:
\begin{equation}
\label{eq:center}
c_{v}^{'(i)} = \frac{\rho}{2} + (1-\rho)c_{v}^{(i)},
\end{equation}
where $\rho\in(0,1)$ is the target activation rate defined below.
This rescaling, together with the way $\sigma_v$ will be defined, ensures that even the extreme centers (at the corners of the hypercube) keep their Gaussian support inside $[0,1]^V$.

The field is then evaluated on a regular grid to translate the pattern for the network.
Note that this grid can differ from the grid for truncated Gaussian center locations.
Let this index set and its generic member be defined as:
\begin{equation}
    \label{eq:Ks}
    \mathcal{K}_s = \prod_{i=1}^{V} \left\{ 1, \dots,m_s^{(i)} \right\}, \ \ \mathbf{k}_s \in \mathcal{K}_s,
\end{equation}
its coordinate along the $i$-th axis, uniformly spaced in $[0,1]$, is defined as:
\begin{equation}
u_{s}^{(i)}(\mathbf{k}_s) = \frac{k_{s}^{(i)} - 1}{m_{s}^{(i)} - 1}, \qquad i = 1,\dots,V.
\end{equation}
Given a center $\mathbf{c}'\in[0,1]^V$ and a standard deviation $\sigma_v>0$, the noise intensity at node $\mathbf{k}_s$ is
\begin{equation}
\label{eq:noise_field}
\nu(\mathbf{c}',\mathbf{k}_s)
= \begin{cases}
\displaystyle
\exp\!\left(-\frac{1}{2}\sum_{i=1}^V \left(\frac{u_v^{(i)}-c_{v}^{'(i)}}{\sigma_v}\right)^{\!2}\right)
  & \text{if this value } \ge \theta,\\[6pt]
0 & \text{otherwise}.
\end{cases}
\end{equation}
The truncation threshold $\theta\in(0,1)$ sets values below it to zero, so that only a localized region of the grid is active; in this implementation, $\theta=0.1$.
The standard deviation $\sigma_V$ is chosen so that a desired rate of nodes is active.
Let $\rho \in (0,1)$ be the target rate of active nodes, i.e., those for which $\nu(\mathbf{c}',\mathbf{k}_s) > 0$.
Denoting by $C_v = \pi^{V/2}/\Gamma(V/2+1)$ the volume of the unit ball in $V$ dimensions, and letting
\begin{equation}
r = \left(\frac{\rho}{C_v}\right)^{1/V}
\end{equation}
be the radius of a ball whose volume equals $\rho$.
We require the Gaussian to drop to $\theta$ at a distance $r$, which gives
\begin{equation}
\sigma_v = \sqrt{\frac{-r^2}{2\ln\theta}}=\left(\frac{\rho}{C_v}\right)^{1/V} \frac{1}{\sqrt{-2\ln\theta}}.
\end{equation}
The parameter $\rho$ represents the proportion of the total volume occupied by non-zero noise regions; a smaller value of $\rho$ generates more exclusive, isolated noise regions, whereas a larger $\rho$ induces significant spatial overlap. Throughout the remainder of this paper, we set $\rho = 0.5$.

A noise pattern in the virtual noise field is expressed as a $V$-dimensional array $\nu(\mathbf{c}',\mathbf{k}_s)$, which must finally be mapped to the network noise field $\mathcal{P}$.
Since each noise pattern activates a sub-region of the network, we must ensure that this sub-region contains a path from the input to the output neurons; otherwise, learning would be impossible.
To guarantee this, we construct the network as a set of fully connected layers in which each hidden layer has $N_s=|\mathcal{K}_s|$ neurons, resulting in $N_h=(L-1)N_s$ hidden neurons in total, and we apply the same noise pattern to every layer, i.e. $p^{(l_1,i)}=p^{(l_2,i)}$ for $1\leq l_1,l_2<L$ and $1\leq i\leq N_s$.
The array is then flattened in standard row-major order into a one-dimensional vector, which is reused identically for all hidden layers.

\begin{figure}[ht!]
\centering

  \begin{subfigure}[b]{\linewidth}
    \centering
    \includegraphics[width=0.33\linewidth]{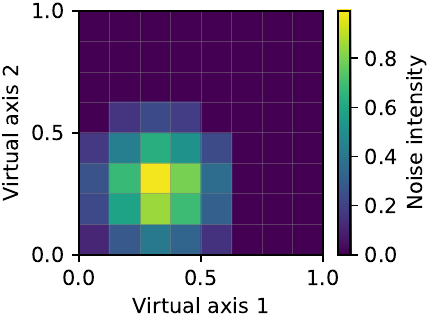}
    \hfill
     \includegraphics[width=0.58\linewidth]{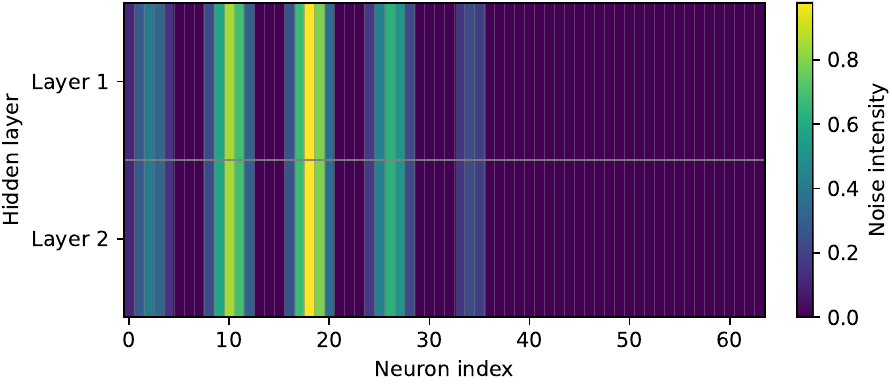}
    \label{fig:pat28}
  \end{subfigure}
  \hfill
  \begin{subfigure}[b]{\linewidth}
    \centering
    \includegraphics[width=0.33\linewidth]{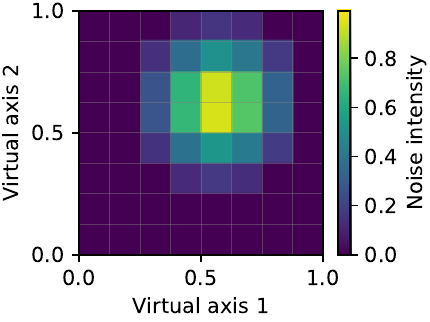}
    \hfill
     \includegraphics[width=0.58\linewidth]{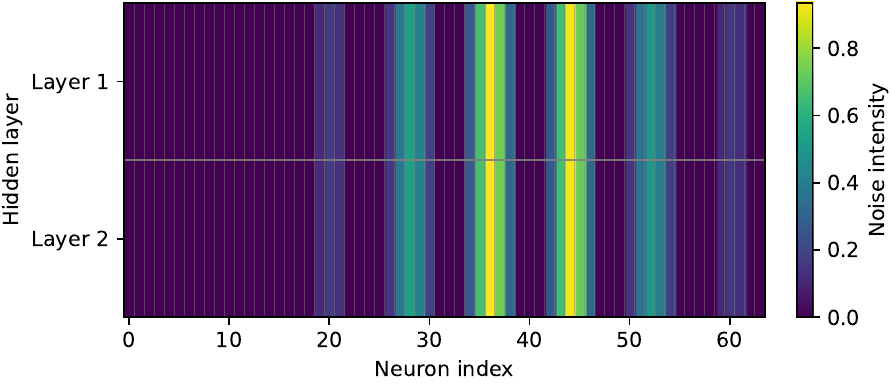}
    \label{fig:pat93}
  \end{subfigure}
  \hfill
\caption{Examples of different noise fields generated via the virtual noise field. Truncated Gaussian placements and noise intensity measurements are given on $\mathcal{K}_v= \{1,\dots,10\}^2$, and $\mathcal{K}_s= \{1,\dots,8\}^2$ regular grids, respectively. (Top) A truncated Gaussian is placed at $(1,2)$. (Bottom) A truncated Gaussian is placed at $(8,7)$. (Left) Noise intensity patterns in the virtual noise field. (Right) Corresponding noise intensity patterns in the network noise field.}
\label{fig:noisefield_ex}
\end{figure}

Figure~\ref{fig:noisefield_ex} shows examples of virtual noise fields and network noise fields.
The heatmap on the left represents the noise intensity distribution (the distribution of the variance of Gaussian noise) in the virtual noise field. 
The virtual noise field is composed of $V=2$, $\mathcal{K}_v= \{1,\dots,10\}^2$, and $\mathcal{K}_s= \{1,\dots,8\}^2$, which means that a two-dimensional neighborhood relationship is introduced for each hidden layer of the network.
The heatmap on the right depicts the corresponding network noise field.
In this case, the network has two hidden layers, each with 64 units.
As mentioned above, the same pattern is applied to each hidden layer, and it can be confirmed that a path for information transmission between the input and output layers is established.
In a fully connected layer, if the order of the weight parameters is changed to match changes in the order of the units, a mathematically equivalent mapping can be maintained before and after the change.
Therefore, the periodically decaying patterns shown on the right side of the figure are generated by the noise field generation process and do not represent five independent pipe-like paths formed from the input to the output.
In other words, interestingly, it can be argued that it is the virtual noise field, rather than the network noise field, that represents physical proximity relationships.

\subsection{Results}
\label{subsec:resultmultifunc}

To verify that a single noise-modulated neural network can learn different functions by using different noise fields, we conduct experiments to approximate various 1D curves.
The target function is defined by $y = \sin(\alpha x+\beta)$ with different parameters $\alpha$ and $\beta$, and is evaluated over the domain $x \in [-2\pi, 2\pi]$. The dataset consists of $N = 1000$ data points uniformly spaced.
Network parameters are optimized by Adam (learning rate $10^{-4}$) with MSE loss.
The maximum number of epochs is $10{,}000$.
Within each epoch, every function (together with its associated noise pattern) is presented to the network exactly once; the order in which functions are presented is randomized at each epoch.

\begin{figure}[ht!]
\centering
    \includegraphics[width=0.95\linewidth]{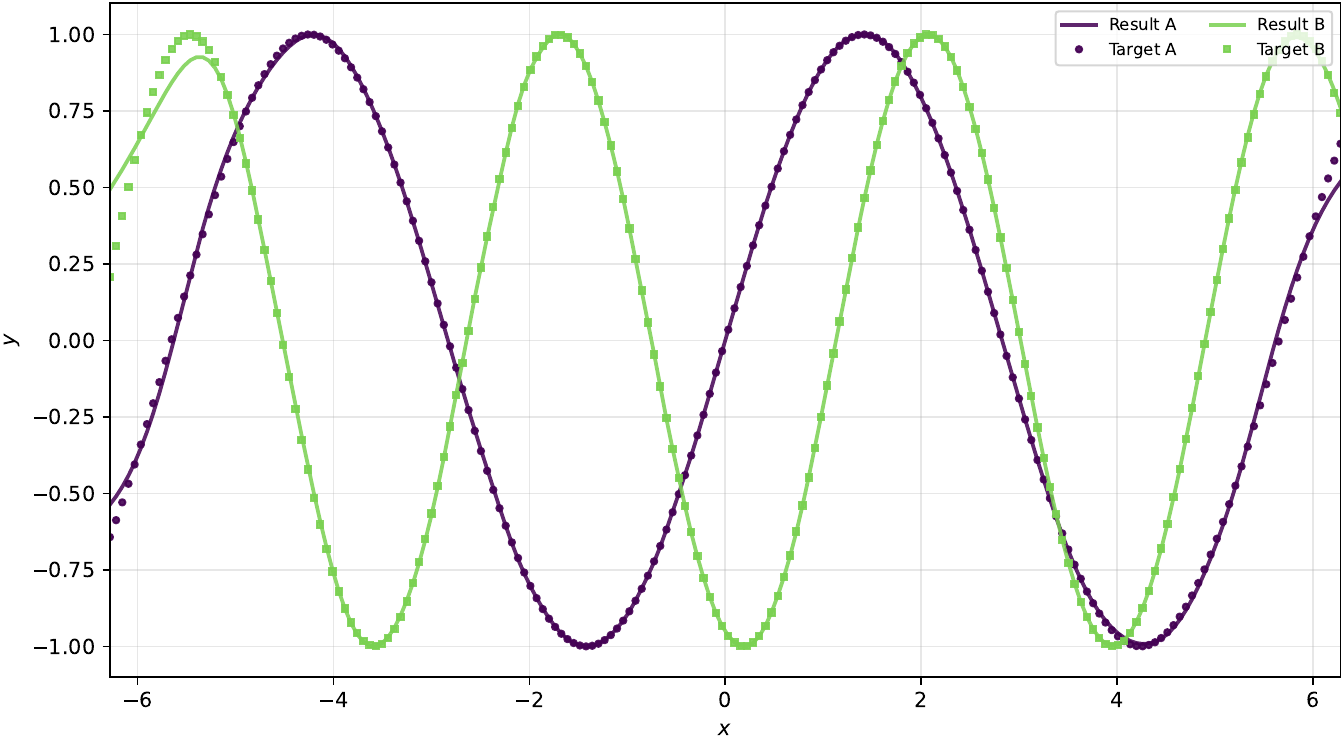 }
   \caption{Inference results (solid lines) compared against the target analytical functions for a single network trained simultaneously on two distinct tasks. By switching the active network noise field between the two patterns shown in Figure~\ref{fig:noisefield_ex}, the network's shared set of trainable parameters successfully alters its output mapping from $y = \sin(1.11x)$ to $y = \sin(1.67x + 4.4)$.}
\label{fig:regression_ex}
\end{figure}

Figure ~\ref{fig:regression_ex} shows the results of training a single noise-modulated neural network to learn two different functions by applying two distinct noise fields, specifically the noise fields shown in Fig.~\ref{fig:noisefield_ex}.
The target functions are $y = \sin(1.11 x+0)$   for the first noise field, and $\sin(1.67 x+4.4)$ for the second noise field.

This example demonstrates that by applying different noise fields, it is possible to train a single Noise-modulated Neural Network to learn multiple target functions. 
Here, if there is no overlap in the non-zero regions of the two noise fields, this is equivalent to training two completely separate Noise-modulated Neural Networks. 
However, the two noise fields used in these results do have overlap in their non-zero regions. 
This implies that components common to the two different target functions are retained in the overlapping regions.

\begin{figure}[ht!]
\centering
    \includegraphics[width=0.5\linewidth]{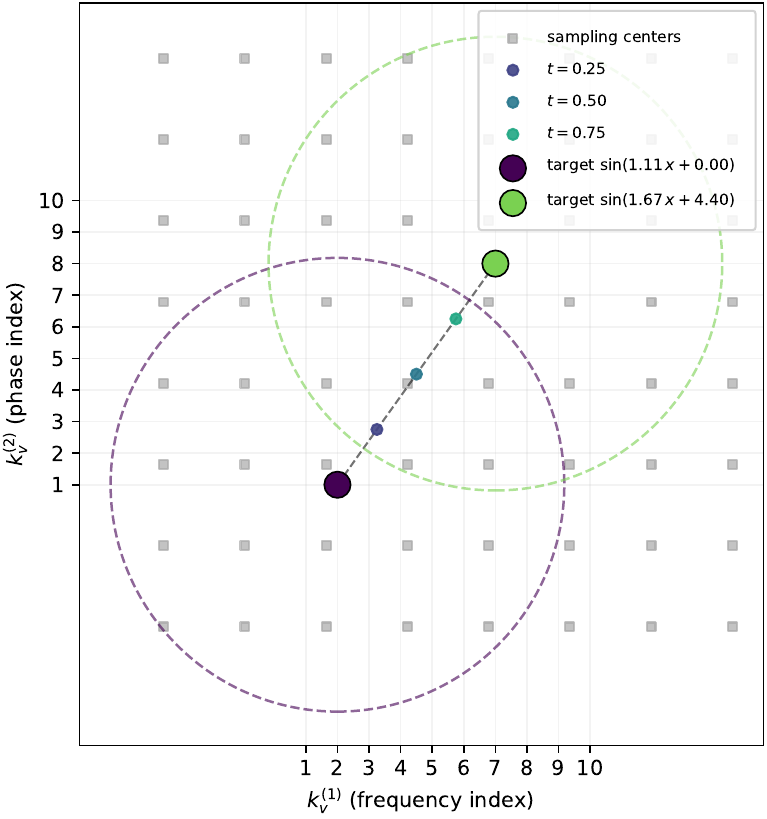 }
\caption{Interpolation of noise fields by moving the center of the activation gaussian. Big dots indicate centers used during training. Small dots indicate gaussian centers used during inference. Dashed circles represent the extremes at which Gaussian are cropped. Squares indicate the position at which the noise intensity value for the actual neurons is sampled.}
\label{fig:trans_concept}
\end{figure}

\begin{figure}[ht!]
 \begin{subfigure}[b]{\linewidth}
    \centering
    \includegraphics[width=0.33\linewidth]{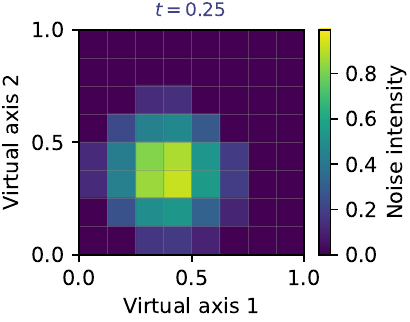}
    \hfill
     \includegraphics[width=0.58\linewidth]{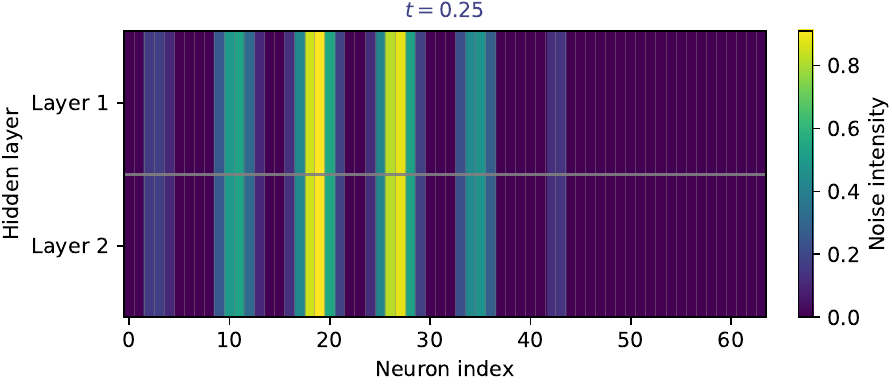}
    \label{fig:pat025}
  \end{subfigure}
  \hfill
\begin{subfigure}[b]{\linewidth}
    \centering
    \includegraphics[width=0.33\linewidth]{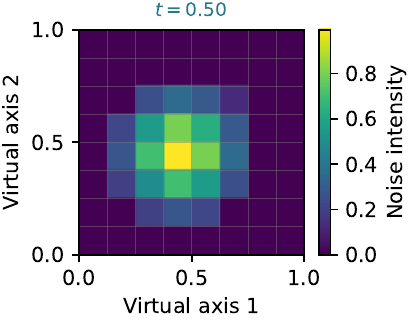}
    \hfill
     \includegraphics[width=0.58\linewidth]{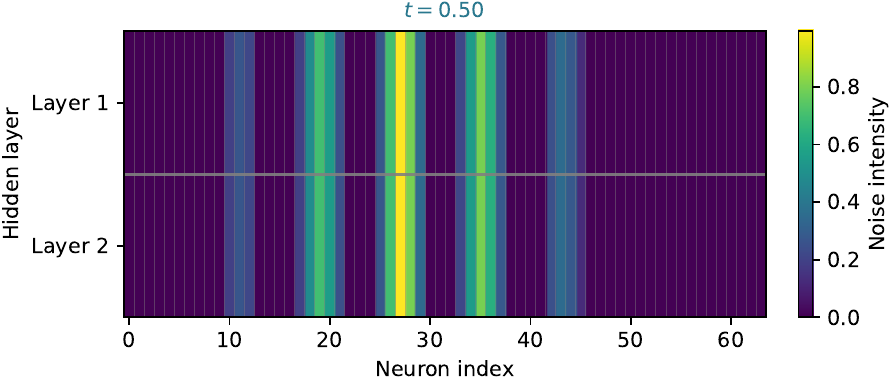}
    \label{fig:pat025}
  \end{subfigure}
  \hfill
  \begin{subfigure}[b]{\linewidth}
    \centering
    \includegraphics[width=0.33\linewidth]{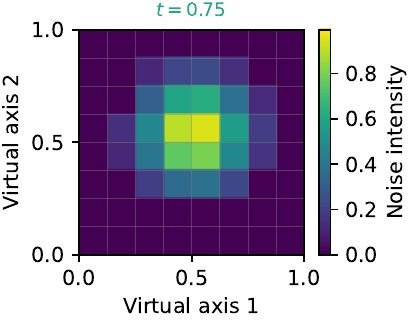}
    \hfill
     \includegraphics[width=0.58\linewidth]{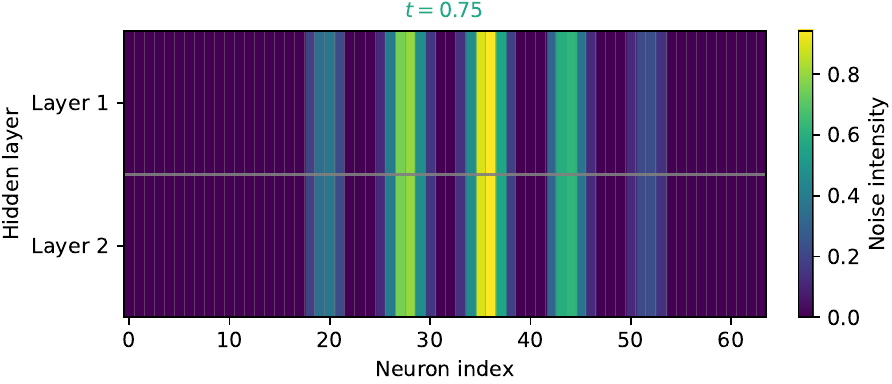}
    \label{fig:pat025}
  \end{subfigure}
  \hfill
  
\caption{Intermediate noise patterns generated during inference by linearly interpolating the Gaussian center between two trained positions. The rows correspond to interpolation parameters $t=0.25$, $t=0.50$, and $t=0.75$, respectively. The left column displays the noise intensity in the virtual noise field, while the right column shows the noise intensity applied to each hidden layer neuron.}
\label{fig:trans_midpattern}
\end{figure}

\begin{figure}[ht!]
\centering
    \includegraphics[width=0.8\linewidth]{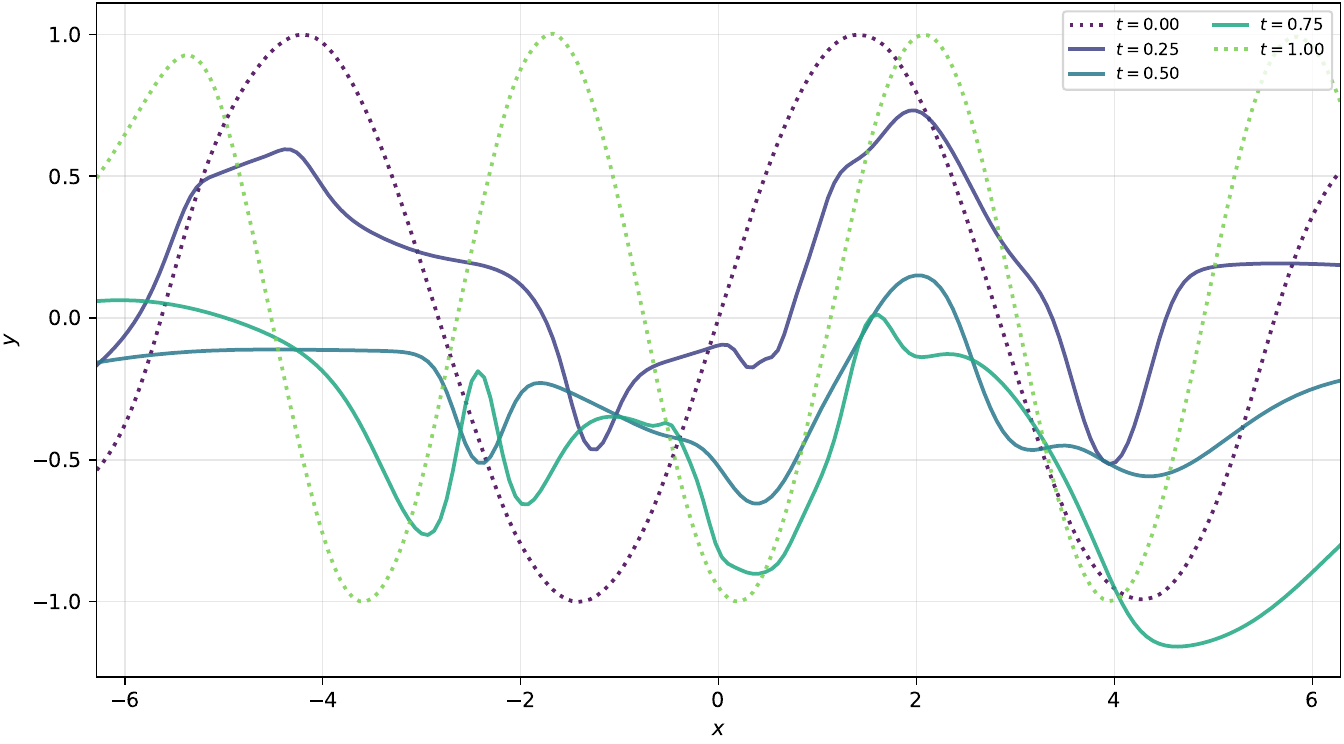}
  \caption{Network inference results observed during the continuous interpolation of the noise field. Inference for trained noise fields are shown in dotted lines, and correspond to the results of Fig.~\ref{fig:regression_ex}. Other inference results, obtained as the parameter $t$ transitions from 0 (first trained function) to 1 (second trained function), are shown with solid lines. }
\label{fig:trans_regression}
\end{figure}

It is interesting to investigate what happens if, during inference, the noise field is varied smoothly between the two noise fields used during training.
Fig.~\ref{fig:trans_concept} provides a schematic representation of this process.
We can consider noise fields whose centers lie over the interpolating line of the two trained functions.
Let us parametrize these centers by $t$, with $t=0$ corresponding to the first noise field and $t=1$ corresponding to the second noise field.
Figure~\ref{fig:trans_midpattern} shows the activations obtained in this way.
Figure~\ref{fig:trans_regression} shows the mapping obtained by linearly interpolating the two noise fields.
The ability to consider the proximity and interpolation between noise fields, regardless of the network structure or the complexity of the noise field patterns, is an advantage of the virtual noise field.

These results show that even when there are overlapping non-zero regions in the noise fields, the interpolation does not necessarily result in a smooth outcome.
It can also be inferred that the complex functions resulting from the interpolation implicitly indicate that there is still room for learning in this region.
Investigating how many functions can be learned within a continuous interval of the virtual noise field and whether this relates to its memory capacity and the arrangement of functions is considered a key focus in advancing a constructivist understanding of the brain using Noise-modulated Neural Networks.
We will focus on this point in the next section.

\section{Memory capacity in spatial partial functionalization}
\label{sec:memory}

In this section, we investigate how many different functions a single noise-modulated neural network can store when each function is associated with a distinct, spatially localized noise pattern.
The procedure involves three stages that pass through several distinct representations.
In the first stage, we generate a family of target functions, each identified by a point in a
$F$-dimensional function parameter space.
This will be detailed in Section~\ref{subsec:functionSet}.
In the second stage, we map every function to a position in a $V$-dimensional virtual noise field.
This will be detailed in Section~\ref{subsec:assignment}.
In the third stage, the network is trained on the entire function family, as described in Section~\ref{subsec:procedure}.
Finally, Section~\ref{subsec:evaluation} presents the evaluation results and the discussion on memory capacity.

\subsection{Function set}
\label{subsec:functionSet}

The first step is to construct the set of target functions that the network must store.
To do this, we first define an $F$-dimensional parameter space. 
For each setting of the parameters, a different function can therefore be obtained. 
To obtain a set of functions, we generate a $F$-dimensional uniformly spaced lattice whose index set and a member are defined as follows:
\begin{equation}
    \label{eq:Kf}
    \mathcal{K}_f = \prod_{i=1}^{F} \left\{ 1, \dots,m_f^{(i)} \right\}, \ \ \mathbf{k}_f \in \mathcal{K}_f.
\end{equation}
Although the size of this set is $N_f = |\mathcal{K}_f|$, the objective here is to first specify $N_f$ and then construct a lattice $\mathcal{K}_f$ with that number of elements.
Since multiple factorizations of $m_{f}^{(i)}$ can yield the same $N_f$, we uniquely define the grid topology by selecting the tuple $(m_{f}^{(i)},m_{f}^{(j)},\ldots m_{f}^{(F)})$ such that $m_{f}^{(i)}\leq m_{f}^{(j)} \; \forall i<j$ and such that $(|m_{f}^{(F)}-m_{f}^{(1)}|,|m_{f}^{(F)}-m_{f}^{(2)}|,\ldots,|m_{f}^{(F)}-m_{f}^{(F-1)}|)$ is lexicographically minimal.

In our experiments, we use the family of sine functions with a phase in $[0,2\pi)$ and a frequency in $[1,2]$, i.e., a 2-dimensional parameter space:
\begin{eqnarray}
    f_{\mathbf{k}_f}(x)= \sin\left( \left(1+\frac{k_{f}^{(1)}-1}{m_{f}^{(1)}}\right) \left(x+2\pi\frac{ k_{f}^{(2)}-1}{m_{f}^{(2)}} \right) \right) \label{eq:funcSine}
\end{eqnarray}
where $\mathbf{k}_f = (k_{f}^{(1)}, k_{f}^{(2)})$.

\subsection{Assignment to Virtual Noise Field}
\label{subsec:assignment}

Based on the virtual noise field defined in Section~\ref{subsec:noisefieldgeneration}, we now assign each function to a truncated Gaussian center within it and then map the resulting noise field of the actual network.
The function parameter space is $F$-dimensional, whereas the virtual noise field has a dimensionality of $V$, and these two may differ.
We therefore introduce the following rule that maps every function to a position in the $V$-dimensional placement grid:
\begin{itemize}
    \item When $V=F$, define the placement grid $\mathcal{K}_v$ of the virtual noise field so that it is congruent with the function lattice and map the functions directly.
    \item When $V>F$, the function lattice is mapped to the first $F$ axes of the $V$-dimensional space, assigning coordinate $0$ to the trailing $V-F$ axes.
    \item When $V<F$, the function lattice must be projected onto fewer axes, which we achieve by collapsing its trailing dimensions into the last axis of the noise field.
\end{itemize}

In performing such a collapse, the order in which functions are enumerated along the placement grid matters because it determines which functions become spatial neighbors in the noise field.
We therefore use a boustrophedon (serpentine) traversal of the $F$-dimensional placement grid, in which the last axis ($k_{f}^{(V)}$) changes fastest. 
Whenever the index along axis $i$ reaches a boundary ($k_{f}^{(i)}=1$ or $k_{f}^{(i)}=m_{f}^{(i)}$), its direction is reversed while the next-slower axis ($k_{f}^{(i-1)}$) advances by one step.
This is the multi-dimensional generalization of the reflected Gray code.
Every successive grid position differs from its predecessor in exactly one axis and by exactly one step, so consecutive functions are always nearest neighbors on the grid and their Gaussian noise fields overlap maximally.
For example, for a $2\times 4$ grid, the traversal visits $(1,1),\;(1,2),\;(1,3),\;(1,4),\;(2,4),\;(2,3),\;(2,2),\;(2,1)$, and the second index sweeps forward along the first row, then backward along the second.
The same serpentine order is applied to both the original function lattice and the $V$-dimensional placement grid, so that functions that are neighbors in the original parameter space remain neighbors along the collapsed axis, preserving local continuity as much as possible.
Each function index $(k_{f}^{(1)},\ldots,k_{f}^{(F)})$ is thereby converted into a grid index $(k_{v}^{(1)},\ldots, k_{v}^{(V)})$, which fixes that function's Gaussian center through Eqs.~(\ref{eq:rawcenter}) and~(\ref{eq:center}).

\subsection{Training settings}
\label{subsec:procedure}

To evaluate memory capacity, we prepare the network as a fully connected feedforward architecture with two hidden layers.
Each hidden layer contains 64 units, and this number remains fixed for all subsequent conditions.
Network parameters are optimized by Adam (learning rate $10^{-4}$) with MSE loss.
Within each epoch, every function (together with its associated noise pattern) is presented to the network exactly once; the order in which functions are presented is randomized at each epoch.

Early stopping is used to terminate training when progress saturates.
Every $\Delta_e$ epochs, the summed training loss over all functions, $S_t$, is evaluated.
Let $S^\star_t = \min_{\tau\le t} S_\tau$ be the best value observed so far.
An update is counted as having no significant improvement when
\begin{equation}
S_t > S^\star_t - \epsilon.
\end{equation}
Training stops after $P$ consecutive no-improvement checks.
In this implementation, $(\Delta_e,\epsilon,P)=(50,10^{-3},10)$ is employed.
In the case where the early stopping condition has not been met, training stops at the maximum number of epochs that is set to $10{,}000$.

\subsection{Evaluation results}
\label{subsec:evaluation}

Given Eq.~\ref{eq:funcSine}, we study three cases. 
Two unidimensional function spaces, where only the phase ($k_f^{(2)}$) or the frequency ($k_f^{(1)}$) is changed (and the other parameter is left at 1), and one bidimensional function space where both indices are varied. 
In this latter case, we also study what happens for both $V=2$ and $V=1$.
For each case, we compare two placement conditions:
\begin{itemize}
  \item \textbf{Ordered}: functions are assigned to noise-field grid positions using the serpentine traversal described in Section~\ref{subsec:assignment}.
  \item \textbf{Shuffled}: the same set of functions is randomly permuted before being assigned to the same grid positions.
\end{itemize}
In the ordered condition, functions that are close in parameter space
occupy nearby grid positions and thus share a large fraction of active neurons.
In the shuffled condition, this locality is deliberately destroyed.

\begin{figure}[h!]
  \centering
  \includegraphics[width=0.72\linewidth]{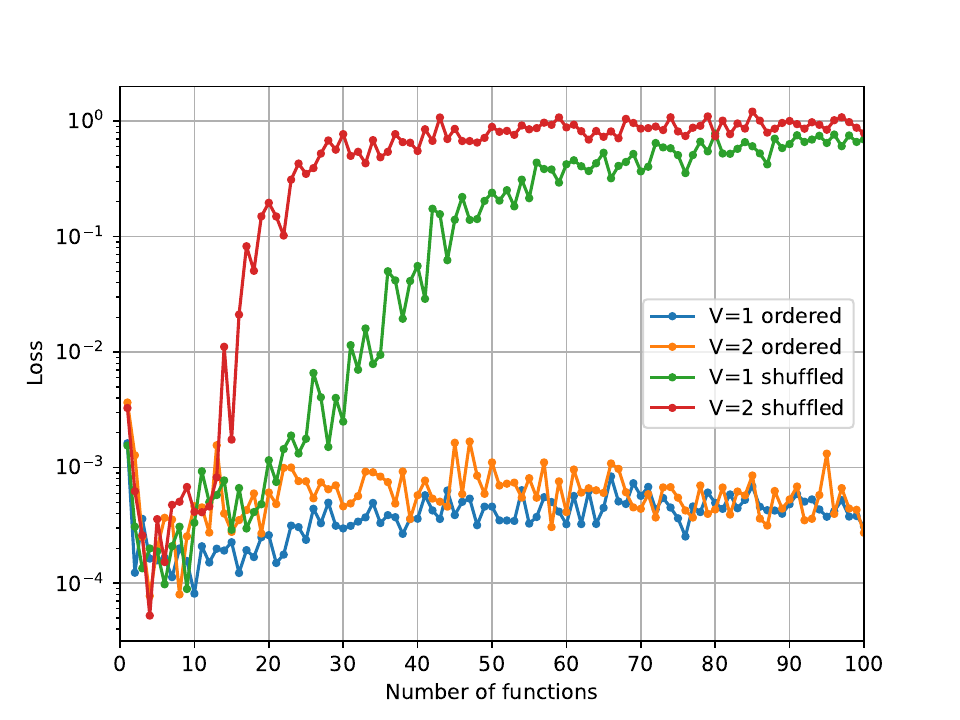}
  \caption{Reconstruction loss for a unidimensional function space with phase variations only, evaluated using unidimensional and bidimensional noise fields. For each condition, functions are assigned either in order or to randomly selected truncated Gaussian centers in the virtual noise field.}
  \label{fig:multidim_fit1}
\end{figure}

\begin{figure}[h!]
  \centering
  \includegraphics[width=0.72\linewidth]{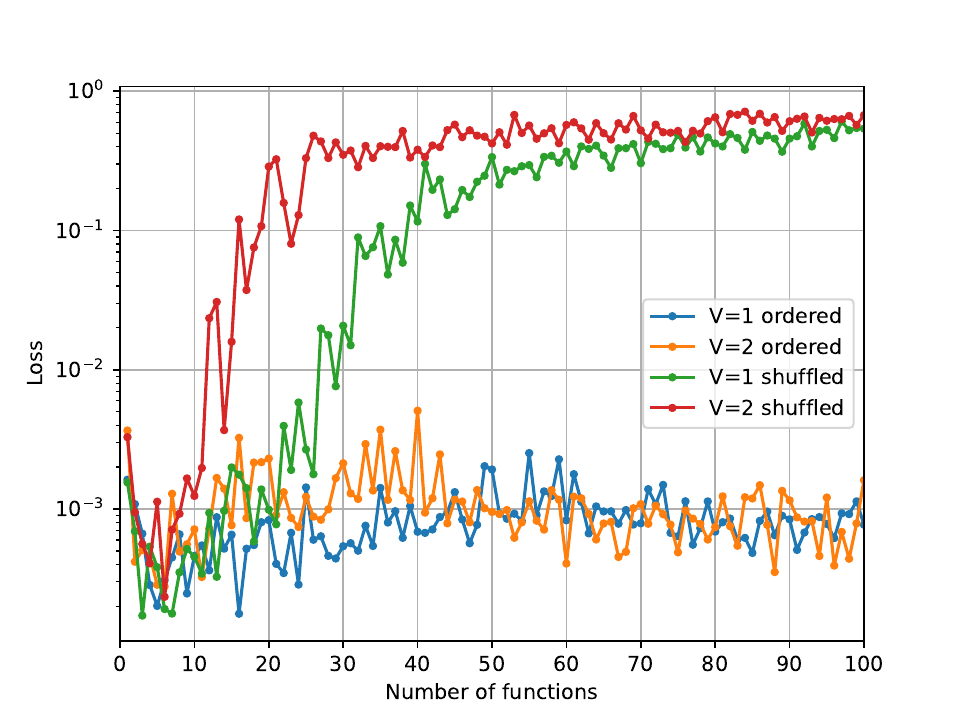}
  \caption{Reconstruction loss for a unidimensional function space with frequency variations only, evaluated using unidimensional and bidimensional noise fields. For each condition, functions are assigned either in order or to randomly selected truncated Gaussian centers in the virtual noise field.}
  \label{fig:multidim_fit2}
\end{figure}

\begin{figure}[h!]
  \centering
  \includegraphics[width=0.72\linewidth]{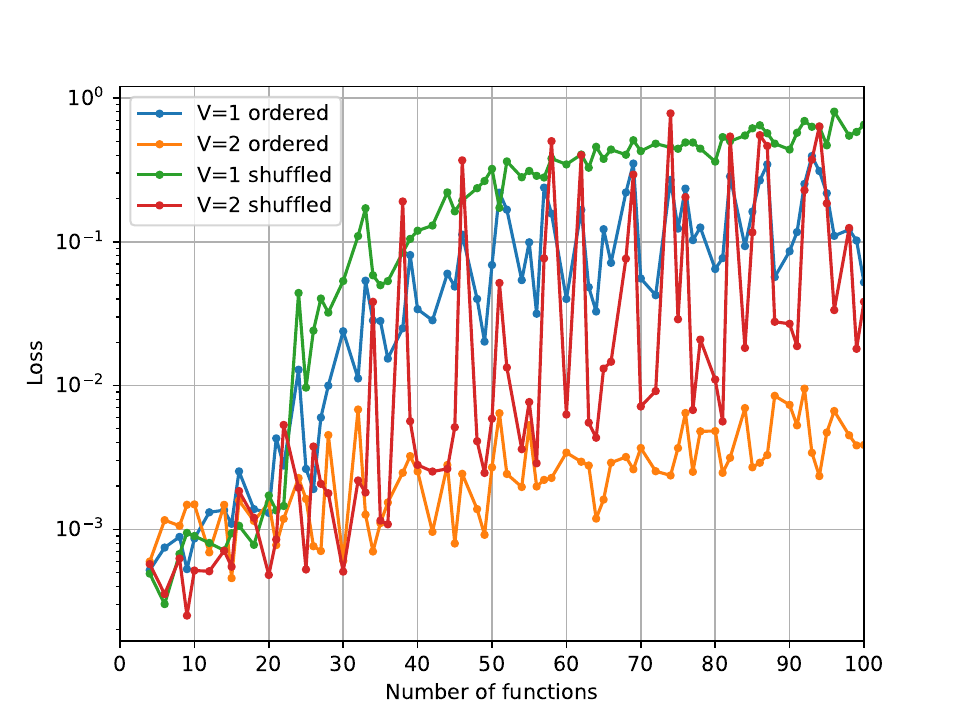}
  \caption{Reconstruction loss for a bidimensional function space with both phase and frequency variations, evaluated using unidimensional and bidimensional noise fields. For each condition, functions are assigned either in order or to randomly selected truncated Gaussian centers in the virtual noise field.}
  \label{fig:multidim_fit3}
\end{figure}

After training, the loss for a family of $N_f$ functions is evaluated as
\begin{equation}
\mathcal{L}_{\max} = \max_{\mathbf{k}_f} \; \mathcal{L}_{\mathbf{k}_f},
\end{equation}
where $\mathcal{L}_{\mathbf{k_f}}$ is the MSE for function $f_{\mathbf{k_f}}$.
Intuitively, the lower this quantity is, the better the network is able to store all the functions at once. We therefore use this worst-case reconstruction error as an inverse proxy for memory capacity.
The curves of $\mathcal{L}_{\max}$ versus $N_f$ for ordered and shuffled conditions for the unidimensional function spaces are shown in Figure~\ref{fig:multidim_fit1} and~\ref{fig:multidim_fit2}. The bidimensional case is shown in Figure~\ref{fig:multidim_fit3}.

Overall, these results suggest that the network’s memory capacity can be improved by introducing a noise field whose spatial structure reflects the similarity properties of the functions being learned.
In Figure ~\ref{fig:multidim_fit1}, since the training data consists of a set of functions where only the phase has been varied, their proximity relationships can be represented in one dimension.
The $V=1$ ordered condition, in which functions with similar phases are placed next to each other, most accurately captures these proximity relationships.
Consequently, even when training on 100 different functions, there is no increase in loss that would indicate a limit to memory capacity.
However, the $V=2$ Ordered condition, which introduces redundant representations of proximity relationships, also achieves a comparable memory capacity, and at this stage, no difference between $V=1$ and $V=2$ can be observed.

The difference becomes apparent under the Shuffled condition.
Under the $V=2$ Shuffled condition, the loss begins to increase sharply once the number of distinct functions reaches around 10, and memory capacity is almost completely exhausted by around 30 functions.
In contrast, under the $V=1$ Shuffled condition, the loss begins to increase gradually once the number of distinct functions reaches around 20, and memory capacity is almost completely exhausted by around 60 functions.
This degradation of the performance can be explained as follows.
First, under the Shuffled condition, the amount of information that can be learned in common by the two functions decreases in the overlapping non-zero regions between the two noise fields.
Consequently, these overlapping non-zero regions cannot be effectively utilized, and the decrease in the “occupied region” per function as the number of functions increases results in an increase in loss.

In this case, if $V=1$, only the two neighboring functions share large overlapping non-zero regions, but if $V=2$, the four neighboring functions must share large overlapping non-zero regions.
Consequently, as the number of functions to be learned increases, the problem of reduced occupied region progresses much more rapidly for $V=2$ compared to $V=1$, resulting in memory capacity being exhausted early on.
This trend is also observed in Figure~\ref{fig:multidim_fit2}, which uses a function set where only the frequency is varied as the training target, supporting the claim that accurately capturing the proximity relationships within the function set through the noise field increases memory capacity.

Figure ~\ref{fig:multidim_fit3} provides further insight into this. 
Here, the function set being learned varies not only in phase but also in frequency, requiring two dimensions to represent its proximity relationships.
Therefore, the $V=2$ Ordered condition is best suited to represent the proximity relationships within this function set in the noise field, whereas the $V=1$ Shuffled condition is the least effective. In fact, the former yields the largest memory capacity, while the latter yields the smallest.

Notably, a comparison between the $V=1$ Ordered condition and the $V=2$ Shuffled condition is of particular interest.
In the $V=1$ Ordered condition, although the dimension of the noise field is insufficient to represent the proximity relationships of the function set, the use of a serpentine arrangement suggests that there is a certain degree of similarity between adjacent functions within that one-dimensional proximity relationship.
On the other hand, under the $V=2$ Shuffled condition, although a noise field of the same dimension as the proximity relationships in the function set is adopted, the shuffling of the function arrangement should destroy the similarity with actually adjacent functions.
Since whether overlapping non-zero regions can be effectively exploited depends on the similarity to neighboring functions, one might expect an increase in loss to be observed under the $V=2$ Shuffled condition before the $V=1$ Ordered condition.
However, in reality, the loss is smaller under the $V=2$ Shuffled condition for most numbers of functions, although the $V=2$ Shuffled condition occasionally exhibits a sharp increase in loss at certain numbers of functions, which will be discussed in the next section.
This result suggests that aligning the dimension of the noise field with the dimension of the proximity relationships in the function set yields positive results, regardless of whether the functions are ordered or shuffled.
In other words, it can be said that the extent to which overlapping non-zero regions are exploited depends not only on the similarity of adjacent functions, but rather, to a large extent, on whether the dimensional parameter $V$ corresponds to the complexity of the proximity relationships and similarities present across the entire function set.
Since increasing the parameter $V$, which represents the dimension of the noise field, increases the number of proximity patterns with large overlapping non-zero regions around a given noise pattern, it might intuitively be thought to lead to monotonic performance degradation under the Shuffled condition.
However, the fact that an independent benefit of matching the dimensions of the function set and the noise field was confirmed provides additional support for the hypothesis that “the network’s memory capacity increases by introducing a noise field that reflects the similarity properties of the functions being learned.”

\section{Discussion}
\label{sec:discussion}

\begin{figure}[ht!]
  \centering
  \includegraphics[width=0.75\linewidth]{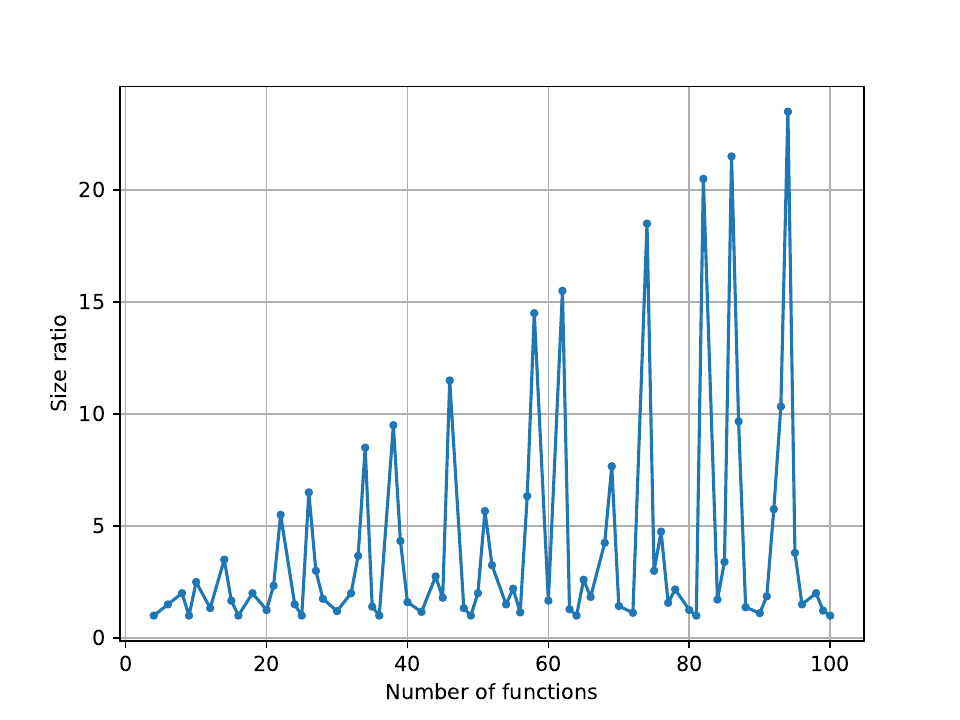}
  \caption{The ratio of grid dimensions, $m_{v}^{(2)}/m_{v}^{(1)}$, plotted as a function of the total number of target functions $N_f$. Periodic spikes represent highly disproportionate grid aspect ratios caused by integer factorization constraints during grid construction. These extreme imbalances effectively collapse the dimensionality of the function space.}
  \label{fig:sizeratio}
\end{figure}

In Figure ~\ref{fig:multidim_fit3}, under the $V=2$ Shuffled condition, a behavior was observed in which the loss increases sharply for a specific number of functions. 
Empirical validation, including variations in early stopping criteria and random seed initializations, confirmed that the specific function counts triggering these loss spikes are highly deterministic.
Therefore, we believe this phenomenon arises from causes embedded within the framework of spatial and partial functionalization, including the assignment of the truncated Gaussian and target functions in the virtual noise field and the transformation into the network noise field.

The most plausible explanation is the difference in spatial resolution between the two axes in the grid space at $F=2$ and $V=2$.
As explained in Section ~\ref{subsec:assignment}, under the condition $V=F$, we define the placement grid $\mathcal{K}_v$ so that it is congruent with the grid used for the function set.
Specifically, the spatial resolution of each axis of $\mathcal{K}_v$ is determined by the generation rule for $N_f$ described in Section ~\ref{subsec:functionSet}.
In this rule, to ensure that the product of $m_v^{(1)}$ and $m_v^{(2)}$ equals the number of functions to be learned, values for $m_v^{(1)}$ and $m_v^{(2)}$ are chosen in lexicographical order when the number of functions is not a perfect square.
Constrained by the necessity of integer factorization, this rule is generally optimal. However, for specific prime or imbalanced composite values of $N_f$, it forces a severe aspect ratio imbalance


Figure ~\ref{fig:sizeratio} shows the change in the ratio $m_v^{(2)}/m_v^{(1)}$ as the number of functions increases.
As shown, at specific numbers of functions, $m_v^{(2)}$ can become extremely large compared to $m_v^{(1)}$, and this disparity increases as the number of functions grows.
As explained in Section ~\ref{subsec:noisefieldgeneration}, we generate the network noise field by placing a single truncated Gaussian in the virtual noise field, with its variance fixed isotropically.
Therefore, when $m_v^{(2)}$ becomes much larger than $m_v^{(1)}$, it becomes difficult to generate differences in the $m_v^{(1)}$ direction, resulting in behavior where the dimension substantially degenerates into only the $m_v^{(2)}$ direction.
In this case, the $V=2$ Shuffled condition imposes more limited learning resources than the $V=1$ Shuffled condition, and simultaneously, the similarity between neighboring functions becomes smaller.
Consequently, it is believed that for those function numbers, the $V=2$ Shuffled condition resulted in a larger loss than any other condition.
The sharp loss increase is therefore interpreted as a limitation of the present grid-construction rule rather than as a contradiction of the main trend. 
This observation suggests that memory-capacity estimates can be sensitive to the discretization of the virtual noise field, especially when the two grid axes become highly imbalanced. 
A more systematic treatment of grid construction, anisotropic field shapes, and adaptive placement rules is left for future work.

\section{Conclusion}
\label{sec:conclusion}

In this paper, we investigated spatial partial functionalization in Noise-modulated Neural Networks using noise fields. Rather than treating noise as a disturbance, the proposed framework uses noise as a mechanism for selecting and organizing active subnetworks within a single model.
The main contributions of this study are as follows. First, we formulated the crossing activation function at three implementation levels, namely Sample-level, Statistical-level, and Analytical-level, and showed through parameter-transfer experiments that these implementations share a compatible parameter space. Second, we introduced the noise field and the virtual noise field as topology-defining mechanisms that enable spatially structured activation of partially overlapping subnetworks. Third, we demonstrated that multiple functions can be stored within a single network and that memory capacity improves when the network noise field is defined by the virtual noise field that reflects the proximity relationship in the function set.

The present study suggests several directions for future work. First, a more extensive investigation of the relationship between memory capacity and noise-field structure is required. Due to computational constraints, this study did not evaluate network sizes or benchmark tasks of sufficient scale to fully exploit three-dimensional virtual noise fields. In addition, many parameters involved in generating network noise fields from virtual noise fields remain to be systematically explored. Scaling up the experiments and conducting a more comprehensive parameter study would strengthen the empirical basis of the noise-field concept and clarify its applicability to the spatial partial utilization of neural networks.

Second, future work should address the adaptive generation of noise fields from training data. In the present study, the number of target functions was assumed to be known in advance, and distinct noise fields were prepared accordingly.
A more generalized framework would invert this paradigm, dynamically inducing noise fields directly from the data.
When the training data contain ambiguity, such as different target outputs for the same input, different noise fields could be induced and used to resolve the ambiguity through spatially partial functionalization of the network. Such a framework would provide further insight into brain-inspired learning mechanisms and may also serve as a basis for practical applications involving multi-valued or context-dependent mappings.

Third, the noise field itself should be considered not merely a fixed parameter but as a dynamical system. In the brain, noise fields are likely to vary over time and may be affected by external fluctuations from sensory receptors, self-generated movements, and other internal and environmental factors. Modeling the noise field as a system that interacts with learning dynamics would open a new research direction in neuromorphic computing.

\section*{Acknowledgements}
This work was supported by the Japan Science and Technology Agency (JST) Fusion Oriented Research for Disruptive Science and Technology (FOREST) Program under Grant JPMJFR242H.
This work was supported by the Japan Society for Promotion of Science (JSPS) KAKENHI Grant Number 25H02619.





\bibliographystyle{elsarticle-num}
\bibliography{reference}







\end{document}